# Benchmarking Akan ASR Models Across Domain-Specific Datasets: A Comparative Evaluation of Performance, Scalability, and Adaptability

Mark Atta Mensah[1], Isaac Wiafe*[1], Akon Ekpezu[2], Justice Kwame Appati[1], Jamal-Deen Abdulai[1], Akosua Nyarkoa Wiafe-Akenten[1], Frank Ernest Yeboah[3], Gifty Odame[4]

[1]Department of Computer Science, University of Ghana, Legon, Ghana
[2]Department of Information Processing Science, University of Oulu, Finland
[3]Expeditors International Inc., Romulus, MI, USA
[4]University of Ghana Health Service, University of Ghana, Legon, Ghana
*iwiafe@ug.edu.gh

**Abstract.** Most existing automatic speech recognition (ASR) research evaluate models using in-domain datasets. However, they seldom evaluate how they generalize across diverse speech contexts. This study addresses this gap by benchmarking seven Akan ASR models built on transformer architectures, such as Whisper and Wav2Vec2, using four Akan speech corpora to determine their performance. These datasets encompass various domains, including culturally relevant image descriptions, informal conversations, biblical scripture readings, and spontaneous financial dialogues. A comparison of the word error rate and character error rate highlighted domain dependency, with models performing optimally only within their training domains, while showing marked accuracy degradation in mismatched scenarios. This study also identified distinct error behaviors between the Whisper and Wav2Vec2 architectures. Whereas fine-tuned Whisper Akan models led to more fluent but potentially misleading transcription errors, Wav2Vec2 produced more obvious yet less interpretable outputs when encountering unfamiliar inputs. This trade-off between readability and transparency in ASR errors should be considered when selecting architectures for low-resource language (LRL) applications. These findings highlight the need for targeted domain adaptation techniques, adaptive routing strategies, and multilingual training frameworks for Akan and other LRLs.

## 1.    Introduction

Recent advancements in machine learning techniques and the increased availability of online textual resources have improved natural language processing (NLP) research, particularly in the field of automatic speech recognition (ASR). However, most of these studies are limited to high-resource languages (HRLs) and seldom consider low-resource languages (LRLs) [1,2]. This is arguably due to the limited availability of large, high-quality speech corpora for training robust ASR models and the substantial costs associated with the development of new resources. However, ensuring that ASR systems are inclusive of all languages is essential for realizing the full potential of NLP technology. Such inclusivity supports the preservation of linguistic and cultural heritage and enables documentation and digital processing of underrepresented languages. This will promote technological fairness and equity by making ASR systems accessible in all languages and by advancing NLP methodologies. Accordingly, some researchers have begun to address LRLs despite their relative lack of resources [3–7].

However, the challenge is that researchers often collect their corpora and use them to fine-tune existing ASR models (e.g., [7,8]). Although this approach is cost-effective, it may suffer from potential biases. The various corpora collected may differ in important dimensions, such as recording conditions (e.g. microphone quality, background noise levels, speaker environments), dialectal and phonetic variation (e.g. regional pronunciations or accents), speaker demographics (e.g. age, gender, sociolects), domains, and annotation standards (e.g. transcription conventions and phonemic vs. orthographic transcriptions). These differences introduce dataset-specific biases [9,10], which may result in model overfitting to the characteristics of the training dataset. Consequently, a model trained on one dataset may fail to perform consistently across other datasets, a phenomenon referred to as limited generalization or overfitting.

Therefore, it is imperative to understand the extent to which different ASR models for LRLs generalize across multiple datasets. This will provide relevant insights into the robustness and transferability of these models. It will also inform researchers about potential dataset-specific biases and provoke strategies for domain adaptation or multi-dataset training to enhance generalization performance. Similar to advancements already observed in high-resource languages, this will facilitate an inclusive and equitable ASR system for LRLs that performs well across diverse speakers and settings.

Accordingly, this study aims to conduct a cross-dataset validation to evaluate the performance of transformer-based Akan ASR models across Akan speech corpora. The Akan language is spoken by most Ghanaians [11], yet it is classified as a low-resource language (LRL). More specifically, this study assesses the performance of existing Akan ASR models that have been fine-tuned using transformer architectures, such as Whisper [10] and XLS-R [12] on publicly available Akan corpora to establish the performance parameters and provide insights into their generalisability.

The remainder of this paper is organized as follows. Section 2 presents an overview of the literature on low-resource languages and Akan speech recognition models. Section 3 describes the methods used, including the specific datasets and models examined. Section 4 presents the results of the study. Section 5 discusses the implications of the findings, the limitations of the study, and future work. Section 6 concludes the study.

## 2.     Related Work

### 2.1     Low Resource Languages and Automatic Speech Recognition

Linguistic databases support both human (e.g. writers, speakers, and translators) and technological (e.g. NLP and ASR) resources to produce culturally sensitive translations. High-resource languages, such as English, French, Spanish, and Chinese, have a substantial amount of speech corpora that are useful in ASR. However, low-resource languages (LRLs), particularly those spoken in Africa, lack sufficient training data required to build accurate ASR systems. LRLs have unique vocabularies and linguistic and grammatical structures. And current ASR architectures lack the cultural nuances and context required to understand them. Ghanaian languages, particularly Akan, exemplify low-resource contexts which pose significant challenges in developing robust automatic speech recognition (ASR) systems.

The inclusion of LRLs in ASR development is crucial as it not only bridges the technological divide between high and low-resource environments but promotes linguistic preservation and broad accessibility of speech technologies across diverse linguistic communities [1,10]. To mitigate these issues, some studies have utilized various approaches to generate speech datasets for LRLs in diverse environments. Ibrahim et al. [4] recorded and collected forty-seven (47) hours of Hausa audio speech in a studio, Georgescu et al. [7] recorded 100 hours of Romanian speech data through an online recording application, and Gulkin et al. [13] collected over four hours of Yoruba audio speech in office settings. For Ghanaian languages, Wiafe et al. [6] used a custom-made mobile app to collect 5000 hours of audio speech recordings in five Ghanaian languages including Akan, Ewe, Dagbani, Dagaare, and Ikposo. The audios were recorded in both indoor and outdoor environments and were spontaneous descriptions of culturally relevant images. Similarly, Asamoah et al. [14] implemented a chatbot on WhatsApp to collect approximately 148 hours of sentences from the Wizard-of-Oz financial service application in Twi and Ga. The Akan Bible dataset [15] comprises professionally recorded formal scripture readings, covering approximately 70% of the Akan Bible. Lastly, the Common Voice Akan dataset [16] contains approximately one hour of crowdsourced informal read speech.

### 2.2     Automatic Speech Recognition Models in Akan

Akan is spoken by approximately 10 million people in Ghana and its neighboring countries [17–19]. It has recently become a focal point in automatic speech recognition (ASR) research due to its linguistic richness and sociocultural significance [8,11]. Initial studies, such as those conducted by Boakye-Yiadom [20], utilized traditional acoustic modeling approaches constrained by limited available speech resources. Although several Akan speech corpora exist, such as the UGSpeechData [6], Common Voice Akan dataset [16], Bible dataset (Lagyamfi Akan) [15], and financial inclusion speech dataset [14], models developed using these datasets inherently reflect biases in recording conditions, dialects, acoustic environments, and speaker demographics, limiting generalization across different speech contexts. Recent advancements utilizing multilingual and self-supervised approaches have sought to mitigate issues of data scarcity by leveraging large-scale pre-trained models trained on multilingual corpora [9]. Table 1 summarises current Akan ASR models, architectures, training datasets, and availability status. For conciseness, pseudo-labels (Models 1–7) are used throughout this manuscript.

Table 1: Existing Akan ASR models

| Model Label | Pseudo-label | Base Architecture | Training Data Source | Availability |
|---|---|---|---|---|
| WhisperSmall Akan[1] | Model 1 | Whisper - small | Lagyamfi/Akan_audio_processed[2], Akan Bible (2 hrs) | Public |
| WhisperLarge Akan-v3[3] | Model 2 | Whisper - large | UGSpeechData (100 hours of Akan) [6] | Private |
| UG_AkanWhisper[4] | Model 3 | Whisper – small | UGSpeechData (100 hours of Akan) [6] | Private |
| Wav2Vec2-Akan-100h-Azunre[5] | Model 4 | wav2vec2-large-xlsr-53 | Common Voice Akan dataset [16] | Public |
| Wav2Vec2-Ashesi-10h-ASR-Africa[6] | Model 5 | wav2vec2-xls-r-300m | Financial inclusion speech dataset [14] | Public |
| UG_AkanWav2Vec[7] | Model 6 | wav2vec2-xls-r-300m | UGSpeechData (100 hours of Akan) [6] | Private |
| Wav2Vec2-Akan-100h-ASR-Africa[8] | Model 7 | wav2vec2-xls-r-300m | Undisclosed 100-hour Akan dataset | Public |

The ASR models presented in Table 1 are based on either the Whisper or Wav2Vec2 architecture. Model 1 used the Whisper-small architecture and was fine-tuned on the Bible dataset (Lagyamfi Akan). Hyperparameters included a learning rate of $5e^{-5}$, a cosine learning scheduler, FP16 (half-precision) training, and a total training duration of approximately two hours. It had a word error rate (WER) of approximately 35%. Model 2 used a Whisper-large variant that contained approximately 1.5 billion parameters. It was fine-tuned on the UGSpeechData dataset [6]. The key training hyperparameters involved a learning rate of approximately $1e^{-5}$, AdamW optimization, gradient checkpointing, FP16 precision, and a training checkpoint at 1800 steps. At checkpoint 1800, it achieved a normalized WER of approximately 27%. Model 3 used the Whisper-small architecture. Similar to model 2, it was fine-tuned on the UGSpeechData dataset [6]. The training hyperparameters included a learning rate of approximately $3e^{-5}$, AdamW optimization, gradient accumulation steps, FP16 precision, and fine-tuning checkpoint at step 6750. It achieved a word error rate (WER) of approximately 29%.

For Wav2Vec2-based models, model 4 was trained using the Common Voice Akan dataset [16], but detailed hyperparameters and exact training configurations are not explicitly disclosed. The model architecture consists of 24 hidden transformer layers with a hidden dimension of 1024, an intermediate layer size of 4096, and 16 attention heads for each layer. The additional configuration included convolutional feature extraction layers with kernels [10, 3, 3, 3, 3, 2, 2], strides [5, 2, 2, 2, 2, 2, 2], stable layer normalization, gradient checkpointing, and a masking probability of 0.05 for time masking in feature augmentation. It achieved a word error rate (WER) of 30%. Model 5 was trained on the financial inclusion speech dataset [14]. The key hyperparameters included a learning rate of approximately $3e^{-4}$, a batch size of 32 (effective batch size of 64 with gradient accumulation), AdamW optimization, linear scheduling, and a warm-up ratio of 0.1. Model 5 reported an approximately 7% in-domain WER. Model 6 was fine-tuned using the UGSpeechData dataset [6]. The model was fine-tuned for approximately five epochs using the AdamW optimizer with a peak learning rate of approximately $8e^{-5}$. The training employed linear scheduling, gradient accumulation, and FP16 precision. Upon final evaluation (step 16,000), Model 6 achieved a WER of 31.45%. Model 7 was trained for 50 epochs on approximately 100 hours of speech data, though details about the exact dataset remain undisclosed. The reported hyperparameters included a learning rate of 0.0003, a batch size of 32 (effective batch size of 64 with gradient accumulation steps of 2), AdamW optimization, linear learning rate scheduling with a warmup ratio of 0.1, and training performed with a fixed random seed of 42. It had a WER of 30%. Collectively, these models represent various architectures, training conditions, and dataset domains.

---

[1] https://huggingface.co/GiftMark/akan-whisper-model

[2] https://huggingface.co/datasets/Lagyamfi/akan_audio_processed

[3] https://huggingface.co/cdli

[4] https://huggingface.co/hci-lab-dcug

[5] https://huggingface.co/azunre/wav2vec2large-xlsr-akan

[6] https://huggingface.co/asr-africa/wav2vec2-xls-r-asheshi-akan-10-hours

[7] https://huggingface.co/hci-lab-dcug

[8] https://huggingface.co/asr-africa/wav2vec2-xls-r-akan-100-hours

### 2.3    Cross-dataset Validation and Low-resource Languages

Despite advancements in ASR technologies, developing robust ASR systems for LRLs remains a challenge, primarily because of limited generalization. Typically, ASR models trained on a specific dataset perform well only under the conditions under which they were trained, including specific accents, recording environments, and content domains [21]. Such limitations result primarily from dataset-specific biases, such as variations in microphone quality, background noise, and speaker demographics [10]. Consequently, these models exhibit substantial accuracy degradation [9,22] when introduced to speech data that differ from the baseline training corpus. Given that domain mismatches significantly impact the performance of ASR systems [21,23], cross-dataset validation is imperative.

Cross-dataset validation which involves testing models on out-of-domain data sources, has become a standard practice in high-resource environments [24,25]. It has been used to evaluate the generalisability of neural network-based models [24–26]. Cross-dataset validation identifies potential model overfitting to specific domain characteristics and provides insights into the real-world performance and generalization capabilities of ASR models. However, it has been underexplored in ASR research on LRLs, particularly Akan. Some existing attempts in high-resource environments have also focused on assessing the generalisics of a single model [25].

This study analyzed the characteristics of diverse publicly available Akan datasets, each representing distinct speech domains, and evaluated the performance of existing Akan ASR models under cross-dataset settings in which a model trained on a specific corpus was evaluated on out-of-domain corpora. This approach aims to evaluate the performance consistency, generalization capacity, and robustness of the selected models beyond their original training conditions. The assessed models included those based on prominent transformer architectures, such as Whisper and Wav2Vec2, which have been fine-tuned on datasets such as informal conversational speech, biblical scripture recitations, spontaneous utterances, and financial dialogues.

## 3. Methodology

### 3.1. Cross-validated Datasets

To systematically evaluate Akan automatic speech recognition (ASR) models across multiple contexts, datasets were selected based on three primary criteria: (1) diversity in speech domains to assess model performance across varied linguistic and acoustic conditions; (2) public availability to enable reproducibility and promote open research practices; and (3) direct relevance to real Akan speech to ensure practical applicability to common usage scenarios. Accordingly, four publicly available Akan speech datasets namely UGSpeechData [6], financial inclusion speech data [14], the Bible dataset (Lagyamfi Akan) [15], and the common voice Akan dataset [16] were selected for this study. They represent speech corpora that have been used to train existing Akan ASR models.

The UGSpeechData corpus is a multilingual dataset comprising five Ghanaian languages. However, this study utilized only the labeled Akan corpus of this dataset which contains 100 hours of recorded utterances of spontaneous speech in Akan contributed by diverse speakers. The audio was originally at 44.1 kHz in MP3 format, and each file spanned between 15 and 30 s. This study utilized a test split subset of 120 utterances that were stratified to maintain gender and age balance (60 females, 60 males; spanning age groups 18–50 years). This balanced split was performed to reduce any bias that may arise from a dominant speaker or gender dependency. The domain of UGSpeechData can be considered generic and not domain-specific, as the utterances cut across descriptions of images from 50 broad categories.

Common Voice is a multilingual audio dataset consisting of paired MP3 recordings and corresponding text transcripts in 60 languages, including Twi (Akan). The Common Voice for Akan (Common Voice Corpus 18.0) contains recorded sentences contributed by nine speakers online via crowdsourcing. The sentences covered a mix of generic content and were less than 10 seconds. This dataset was included to evaluate how the models handle open-domain crowdsourced speech that may include colloquial phrases and varied pronunciations. Because of the small size of this corpus, this study used the entire available test split (or validated set) for evaluation.

The Lagyamfi/Akan corpus is a collection of Twi Bible readings by a male narrator and is available on Hugging Face. The recordings, originally at 48 kHz/24-bit, represent literary read speech with formal vocabulary and stylistic features distinct from everyday language. This study used a 2-hour subset of 259 test passages, unseen during training, to evaluate model performance on scripture domain speech.

The financial Inclusion speech dataset consists of approximately 107 hours of Akan phrases, each less than five seconds long, spoken by participants across three dialects: Asante, Akuapem, and Fante. This study extracted a 10-hour Akan subset from this dataset to assess the ASR model performance on what was presumed to be domain-specific, semi-spontaneous speech. Prior to evaluation, these datasets were standardized to mono-channel audio with a 16 kHz sampling rate and 16-bit PCM WAV encoding. The original training-test splits provided by the dataset authors were maintained consistently throughout the evaluation.

### 3.2. Evaluation Procedure

To ensure computational consistency, the seven models (Models 1–7) presented in Table 1 were evaluated under identical computational conditions using Google Collab with an NVIDIA A100 hardware accelerator. Half precision (FP16) was achieved for all models. To maintain consistency across the models, evaluations were performed using the original dataset splits provided by each source.

Most models were obtained from the Hugging Face Hub. For the private models, the provided checkpoints were used in a controlled environment. All Whisper models were executed using the WhisperForConditionalGeneration API (with greedy or default decoding settings) and all Wav2Vec2 models were executed using Wav2Vec2ForCTC. No external language models were added to the CTC decoding, and the transcripts were raw model outputs. All models were evaluated using the same test sets drawn from each dataset. To ensure consistency, the same audio files were used for each model in a given dataset test portion. Each audio clip was passed on to each model and the transcript output of the model was recorded. Preprocessing was uniform across the models. No model-specific preprocessing was performed beyond what the model's API does internally (for example, whisper models internally segment audio and tokenize, and Wav2Vec2 models operate directly on raw waveform input). The evaluation code processed each utterance sequentially for all models logged the outputs, and ensured identical conditions.

For performance evaluation, two primary metrics were computed: Word Error Rate (WER) and Character Error Rate (CER), using the Jiwer Python library. They represent the most utilized metrics for evaluating the performance of ASR models [27]. To match the reference transcripts and focus solely on the word content, predicted transcriptions were normalized by converting the text to lowercase and removing punctuation and special characters. This is important because some models (Whisper) output punctuation or casing, while others (Wav2Vec2) output lowercase text with no punctuation. No language-specific normalization (e.g. handling of tone marks or diacritics) was applied beyond the given orthography; the reference transcripts themselves used the standard Twi Latin script without diacritics, so this was acceptable. Also, to assess result variability, the WER and CER were calculated for each dataset-model combination, accompanied by the standard deviation and 95% confidence intervals.

### 3.3. Qualitative Error Analysis

In addition to the quantitative metrics, qualitative error analysis on the predicted transcription was conducted. This involved examining examples of transcripts in which models had high errors in identifying common error types. For instance, specific phonemes that were consistently misrecognised or particular words that were out of vocabulary (OOV) for a model. In addition, error patterns across models were compared; for example, did whisper-based models make different kinds of mistakes than Wav2Vec2 models on the same audio? Particular attention was paid to whether an error was due to a dialectal difference (the model outputs a synonym or variant spelling), a function word insertion (such as adding "the" or an extra filler word), or a content word mistake.

Errors were classified into common linguistic categories, such as phoneme-level confusion, out-of-vocabulary terms, dialectal mismatches, insertion or omission of words, and lexical substitutions. This qualitative analysis enabled us to understand why certain models performed better or worse in different scenarios.

## 4.     Results and Findings

Comparative evaluations illustrate significant generalization failures when models trained in one domain are tested in another, underscoring the substantial limitations of current domain-specific training approaches. Generally, models trained on formal speech datasets often struggled with informal conversational speech or dialect-specific vocabulary, whereas those fine-tuned on conversational data demonstrated limited accuracy in formal speech content. In addition, distinct error profiles were observed between the whisper (sequence-to-sequence) and Wav2Vec2 (CTC-based) models. Whisper-based models produced fluent Akan transcriptions, even when inaccurate, potentially facilitating easier downstream text processing. Wav2Vec2-based models often generate fragmented and incomplete outputs, particularly with unfamiliar vocabulary, underscoring their acoustic and linguistic limitations. These outcomes reinforce the fact that robust domain generalization in low-resource Akan ASR remains challenging. Consequently, systematic cross-domain evaluations and targeted domain adaptation strategies are essential for developing reliable ASR systems applicable to Akan and similar LRLs. The following subsections present the results of the model evaluations across the four datasets.

### 4.1.     UGSpeechData

Performance evaluation on the UGSpeechData corpus showed variations in the performance of the evaluated ASR models. Model 3 achieved the best performance, with a WER of approximately 30% and a CER of approximately 12%. This was closely followed by Models 7, 2, and 6, all of which achieved a WER and CER of less than 40% and 15%, respectively. The results for Models 2,3, and 6 reflect in-domain resemblance as they were all originally trained on the UGSpeechData. Fig. 1 presents a comparative analysis of the performance of Models 1 to 7 on the UGSpeechData corpus.

Interestingly, Model 7 which originally had no exposure to the UGSpeechData, performed competitively, particularly in terms of CER. This suggests that its acoustic model generalizes to Akan phonemes, even though its lexicon is not domain-specific. Apart from Model 7, other models trained on unrelated datasets showed poor domain alignment. Models 1 and 4 showed domain mismatch, with a WER and CER of approximately 80% and 30% respectively. Model 5 performed the worst overall, with a WER and CER exceeding 90% and 80%, respectively. This indicates both acoustic and lexical mismatches.

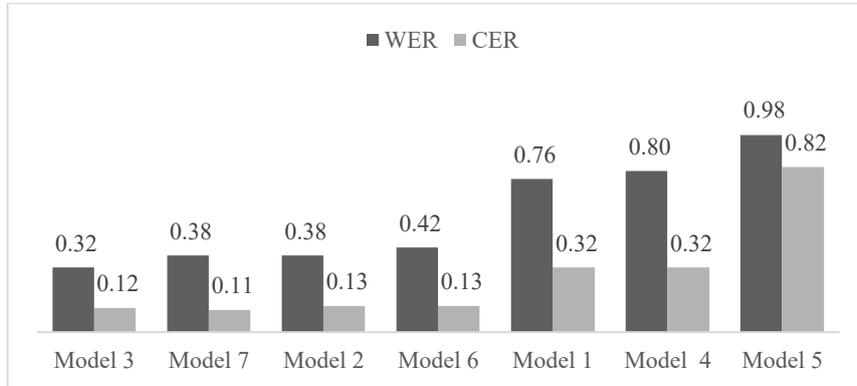

**Fig. 1.** Comparative analysis of model performance on UGSpeechData corpus

These findings reinforce the importance of domain alignment in low-resource ASR systems. In particular, the drop in performance of models trained outside UGSpeechData confirms that narrow-domain ASR models struggle to generalize. Furthermore, the qualitative error analysis showed that frequent errors included substitution of acoustically similar Akan words and omission of vocabulary items unfamiliar to the training domain

### 4.2.    Common Voice Akan (Twi)

The evaluation of ASR models on the Common Voice Akan dataset which consists of short, crowd-sourced utterances contributed by nine speakers, also showed considerable differences in performance. Surprisingly, Model 1 which was originally trained on Akan Bible readings from a single speaker, achieved the lowest performance among the other evaluated models, with a WER of 64% and a CER of 23%. This is notable because the training data are narrow in style; that is, they are scripted and narrative, domain-specific (Bible), and speaker profile (one male speaker). Fig. 2 summarises the comparative performance of all evaluated models.

In contrast, Model 4, which was trained specifically on the Common Voice Akan corpus, performed considerably worse (WER = 82%; CER = 28%). Although domain alignment might have been expected to confer an advantage, the performance gap suggests that other factors may have contributed. However, because the hyperparameters (e.g. batch size, learning rate, and training steps) were not disclosed, it is difficult to assess whether architectural depth, insufficient data variation, or lack of pre-trained initialization may have contributed to the underperformance. This suggests that domain-matched training alignment alone does not guarantee robust generalization without architectural and optimization rigor requirements.

Models 3 and 6 which were both originally trained on the UGSpeechData, demonstrated weak generalization to the Common Voice test set. For instance, Model 3, which performed best on UGSpeechData, deteriorated here (WER: 767%, CER: 29%), whereas Model 6 achieved a lower WER of 69% and CER of 21%. While this may be an indication of good phoneme decoding but mismatches in word segmentation or lexicon, the performance gap between Models 3 and 6 also indicates differences in the transformer architectures. Model 3 was based on Whisper, whereas Model 6 used Wav2Vec 2.0. Wav2Vec2's frame-level feature learning and reduced dependence on language model decoding most likely contributed to its superior resilience in the informal, acoustically variable conditions of Common Voice. Model 2 which was built on the Whisper-large architecture (approximately 1.5 billion parameters), was not evaluated on Common Voice due to configuration constraints that prevented decoding under spontaneous or multi-speaker conditions.

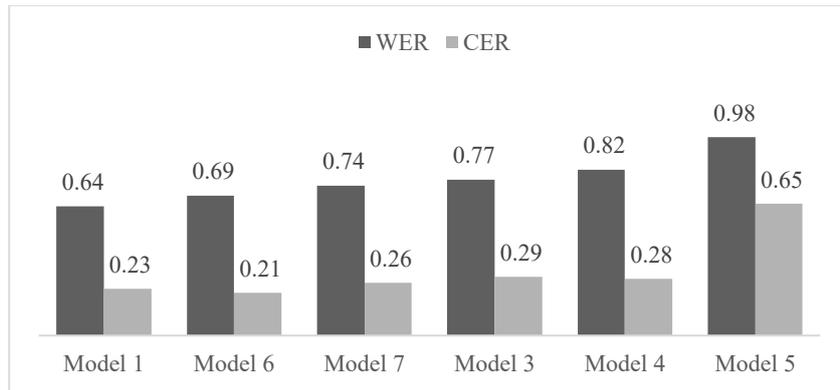

**Fig. 2.** Comparative analysis of model performance on the Common Voice Akan corpus

Model 7 performed comparably to Models 3 and 4 (WER: 74%, CER: 26%), suggesting the moderate transferability of its acoustic representations despite the domain mismatch. Again, Model 5 which had the worst performance with the UGSpeechData corpus, recorded the weakest results (WER: 98%, CER: 65%). This indicates a consistent failure to generalize across all the test domains. Typical recognition errors across the evaluated models involved the misrecognition of informal vocabulary, phoneme confusion, and insertions resulting from acoustic noise.

### 4.3.    Lagyamfi Akan Bible

Model 1, which was originally fine-tuned on this dataset, achieved the lowest WER of approximately 37% and a CER of 11%. Model 4 performed better than the other models and achieved a WER of 47% and a CER of 12%. Its relatively moderate performance may be attributed to its acoustic familiarity with read speech, although it lacked exposure to religious vocabulary and stylistic cues. As shown in Fig. 3, the remaining models show much higher error rates.

Models 6 and 7 had the same WER and CER of approximately 70% and 20%, respectively. Model 3 also performed poorly on this dataset (WER = 78%, CER = 65%). This may indicate a decoder bias when applied to domain-specific vocabulary and stylistic features outside its training scope. Again, Model 5 as with other evaluations, showed the highest degradation, with WER: 95% and CER: 64%. This reflects broad failures at both the acoustic and lexical levels. Model 2 was not evaluated on this dataset due to configuration limitations similar to those affecting Common Voice testing. Across the models, frequent recognition errors involved scripture-specific vocabulary, including archaic terms, formal constructions, and uncommon, biblical names. These often resulted in word substitutions and deletions, particularly for tokens not encountered during training.

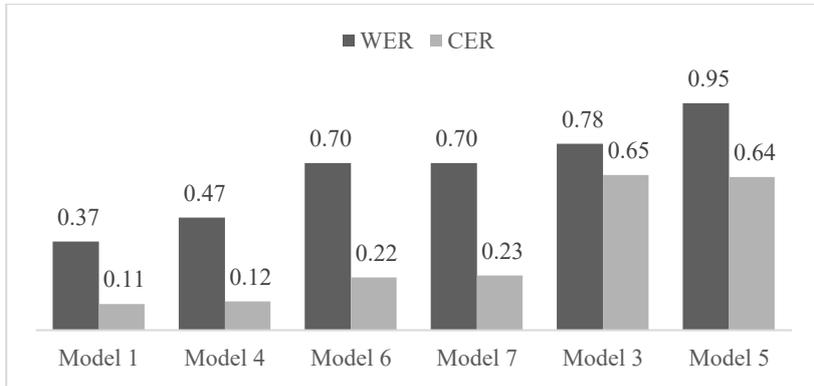

**Fig. 3.** Comparative analysis of model performance on the Akan Bible corpus

### 4.4. Financial Inclusion Speech Dataset

The evaluation of ASR models on the Financial Inclusion Speech corpus exhibited some of the most divergent performances across all datasets, except for the in-domain trained model. Model 5, which was originally trained with this dataset, achieved the lowest WER of approximately 10% and CER of approximately 6%. This reflects a strong domain adaptation. In contrast, all out-of-domain models showed reduced performance. Models 6 and 7 both achieved high WERs of approximately 86%. Model 4 also exhibited a performance mismatch with a WER of 95%. These high error scores indicate a poor fit for the acoustic and linguistic variability in this domain. The comparative performance details are shown in Fig. 4. Model 1 was excluded from this evaluation due to configuration limitations.

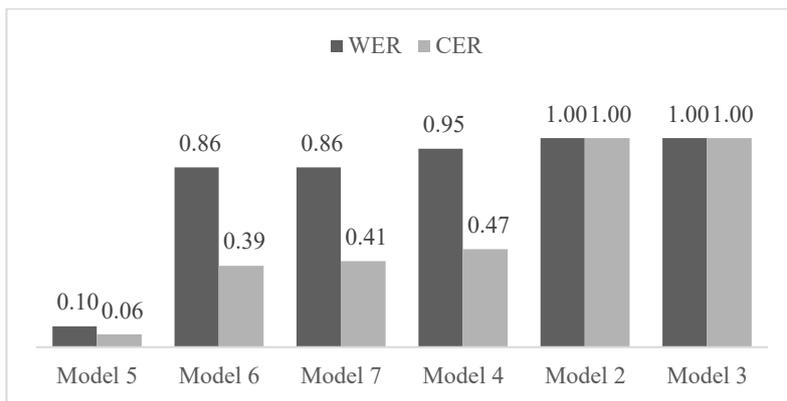

**Fig. 4.** Comparative analysis of model performance on the Financial Inclusion Speech Corpus

Models 2 and 3 which utilized the Whisper architecture encountered the most severe decoding failures; both achieved error rates exceeding 100%. These outcomes are indicative of decoder collapse, in which the model produces long irrelevant sequences or omits large portions of the input. Common recognition errors included the insertion of unrelated words, deletions, and misrecognition of domain-specific terms. These results highlight the inherent challenges of deploying ASR models across significantly different speech domains without targeted adaptation.

## 5.     Discussion

This study evaluated the performance of Whisper- and Wav2Vec-based Akan models across speech datasets representing spontaneous speech (UGSpeechData), informal crowdsourced utterances (Common Voice Akan), religious readings (the Akan Bible), and finance sentences (the Financial Inclusion Speech dataset). Using WER and CER, the study explains how training domain, model architecture, and speech style (formal, informal, spontaneous, and read) influence recognition performance, particularly for LRLs. Ultimately, it affirms ongoing arguments that domain mismatch between training and testing data often leads to a degradation of recognition accuracy [24,28], even in LRLs. The findings also highlight the trade-offs between domain-specific optimization and cross-domain robustness of the model. This emphasizes the need for adaptive ASR strategies tailored to multilingual, low-resource languages.

### 5.1.     Domain-Specific Generalization

The evaluation showed domain dependency among models. Most models consistently achieved lower error rates when evaluated within their original training domains but showed considerable performance degradation on domain-mismatched speech. For example, models fine-tuned on formal biblical speech demonstrated notably higher error rates when processing informal conversational speech, illustrating limitations in cross-domain generalization [29]. Similarly, models optimized for structured formal speech such as the UGSpeechData dataset [6] exhibited performance deterioration when applied to informal financial dialogues. This highlights challenges in effectively adapting models across distinct speech contexts. Thus, Akan ASR models do not inherently generalize across diverse speech domains. Future research should employ targeted domain adaptation strategies to enhance the robustness and applicability of Akan ASR systems.

### 5.2.     Robustness vs. Specialization

Findings from this study revealed a trade-off between domain specialization and general robustness. This supports current knowledge in ASR research [9,10]. Domain-specialized models consistently demonstrate high accuracy within their respective training contexts, outperforming generalized models within those specific domains. For example, the Whisper model fine-tuned on the biblical corpus substantially outperformed its generalized counterparts, achieving a WER of approximately 37% for formal scripture readings. However, this specialized model displayed considerable degradation when evaluated on spontaneous financial dialogues. This indicates overfitting to specific linguistic and acoustic features within its training corpus.

Conversely, models trained on broader mixed-domain datasets exhibited greater resilience across diverse speech conditions. Although these generalized models rarely attain optimal performance in any specific domain, their overall robustness prevents severe recognition failures. An illustrative example is the Wav2Vec2 model (model 6). Despite being trained on UGSpeechData, it demonstrated moderate but stable performance across various speech contexts. This may be attributed to the robust self-supervised acoustic modeling capabilities of Wav2Vec. This finding supports deployment frameworks that strategically combine domain-specialized models, capitalizing on their high domain-specific accuracy, with broadly trained models that provide robust fallback capabilities when processing various speech inputs.

### 5.3. Decoding Bias and Error Interpretability

The results of the qualitative error analysis also highlighted notable decoding differences between sequence-to-sequence models (Whisper) and connectionist temporal classification (CTC)-based (Wav2Vec2) architectures. Prior research has similarly noted architectural biases and their implications for downstream ASR application usability [30]. Whisper's decoding process is influenced by its internal language modeling component; thus, it frequently generates fluent yet semantically incorrect transcriptions. For instance, during the transcription of biblical names, Whisper transformed "Nebukadnezar" into the fluent yet incorrect "Neburankan", preserving phonetic coherence but substantially altering its meaning. In contrast, Wav2Vec2 lacked a strong language modeling bias and thus produced fragmented or incomplete segments, such as "NebukaX". This indicates a decoding failure.

These contrasting error profiles have significant implications for ASR applications in sensitive contexts, such as medical or legal transcription, where subtle inaccuracies potentially pose greater risks than obvious recognition errors. Whisper's fluent but incorrect transcriptions can inadvertently mislead users, whereas the transparently incorrect outputs of Wav2Vec2 provide clearer indicators for manual review and correction of the transcriptions. Consequently, deployment decisions must consider the nature of acceptable error types and their implications for end users, advocating distinct models or decoding strategies tailored specifically to user needs and error tolerance levels.

### 5.4. Limitations and Future Work

This study has several limitations that open avenues for future research. A primary limitation is the variability in the dataset size and quality, which may have introduced biases and limited the validity of architectural comparisons. Future work should employ more controlled experimental setups to isolate the effects of model architecture, data quantity, and corpus quality, as well as explore standardized multilingual training strategies to address these issues.

Another key limitation is the poor handling of Akan-English code-switching, which is a common linguistic feature of Ghanaian speech. Both the Whisper and Wav2Vec2 models failed to process multilingual inputs reliably. They often omitted English segments or inaccurately transcribed them as Akan phonemes, thereby reducing transcription accuracy. Future research should prioritize the development of models that integrate accurate language identification with multilingual transcription capabilities.

Additionally, exploring lightweight domain adaptation techniques, such as adapter modules and continual learning strategies, could enhance model adaptability across speech domains. These approaches may offer practical pathways for building more robust and context-aware ASR systems for low-resource multilingual environments such as Akan.

## 6.     Conclusion

This study presents the first attempt at a cross-dataset evaluation of transformer-based ASR models for the Akan language. By benchmarking seven transformer-based ASR Akan models across four domains, including scripted religious speech to informal and financial dialogues, this study re-emphasizes the challenges of domain mismatch and the limitations of overly specialized systems. Although models such as Whisper and Wav2Vec2 offer distinct advantages, no single approach guarantees optimal performance across all conditions. Thus, researchers and practitioners should consider both domain characteristics and architectural implications when designing or deploying ASR systems for Akan.

The findings of this study advance the existing knowledge of how architectural choices influence error patterns, how domain mismatch affects recognition accuracy, and where current models fall short in multilingual handling. It also sets a precedent for cross-domain evaluation in other low-resource language contexts in the future. This study moves beyond single-dataset evaluations and highlights the real-world challenges of deploying ASR systems in heterogeneous linguistic environments. This supports the broader goal of building inclusive, context-aware speech technologies that can serve underrepresented language communities more effectively.

## Acknowledgment

This study was funded by the "Building a New Generation of Academics in Africa (BANGA-Africa) and Google Gift from Google Research, Ghana.

## References

1.     Besacier L, Barnard E, Karpov A, Schultz T. Automatic speech recognition for under-resourced languages: A survey. Speech Commun 2014;56:85–100.
2.     Rista A, Kadriu A. Automatic Speech Recognition: A Comprehensive Survey. SEEU Review [Internet] 2020;15:86–112. Available from: https://www.sciendo.com/article/10.2478/seeur-2020-0019

3.  Zhuang W, Sun Y. CUTE: A Multilingual Dataset for Enhancing Cross-Lingual Knowledge Transfer in Low-Resource Languages [Internet]. In: Proceedings ofthe 31st International Conference on Computational Linguistics. Abu Dhabi, UAE: Association for Computational Linguistics; 2025 [cited 2025 Jun 23]. page 10037–46.Available from: https://aclanthology.org/2025.coling-main.670/

4.  Ibrahim UA, Boukar MM, Suleiman MA. Development of Hausa dataset a baseline for speech recognition. Data Brief [Internet] 2022;40:107820. Available from: https://doi.org/10.1016/j.dib.2022.107820

5.  Gutkin A, Demirsahin I, Kjartansson O, Rivera C, Túbòsún K. Developing an open-source corpus of yoruba speech. Proceedings of the Annual Conference of the International Speech Communication Association, INTERSPEECH 2020;2020-Octob:404–8.

6.  Wiafe I, Abdulai JD, Ekpezu AO, Helegah RD, Atsakpo ED, Nutrokpor C, et al. UGSpeechData. 2023;

7.  Georgescu AL, Cucu H, Buzo A, Burileanu C. RSC: A romanian read speech corpus for automatic speech recognition. LREC 2020 - 12th International Conference on Language Resources and Evaluation, Conference Proceedings 2020;6606–12.

8.  Salihs SA, Wiafe I, Abdulai JD, Doe Atsakpo E, Ayoka G, Cave R, et al. A Cookbook for Community-driven Data Collection of Impaired Speech in Low-Resource Languages. In: August, editor. Accepted - InterSpeech Conference 2025. 2025.

9.  Baevski A, Zhou H, Mohamed A, Auli M. wav2vec 2.0: A Framework for Self-Supervised Learning of Speech Representations [Internet]. In: Proceedings of the 34th International Conference on Neural Information Processing Systems. Vancouver, BC, Canada: Curran Associates Inc.; 2020. page 12.Available from: https://github.com/pytorch/fairseq

10. Radford A, Kim JW, Xu T, Brockman G, Mcleavey C, Sutskever I. Robust Speech Recognition via Large-Scale Weak Supervision [Internet]. In: ICML'23: Proceedings of the 40th International Conference on Machine Learning. Honolulu, Hawaii, USA: Journal of Machine Learning Research; 2023. page 28492–518.Available from: https://github.com/openai/

11. Agyei E, Zhang X, Bannerman S, Quaye AB, Yussi SB, Agbesi VK. Low resource Twi-English parallel corpus for machine translation in multiple domains (Twi-2-ENG). Discover Computing [Internet] 2024;27:17. Available from: https://link.springer.com/10.1007/s10791-024-09451-8

12. Babu A, Wang C, Tjandra A, Lakhotia K, Xu Q, Goyal N, et al. XLS-R: Self-supervised Cross-lingual Speech Representation Learning at Scale. In: Proceedings of the Annual Conference of the International Speech Communication Association, INTERSPEECH. International Speech Communication Association; 2022. page 2278–82.

13. Gutkin A, Demirsahin I, Kjartansson O, Rivera C, Túbòsún K. Developing an open-source corpus of yoruba speech. Proceedings of the Annual Conference of the International Speech Communication Association, INTERSPEECH 2020;2020-Octob:404–8.

14.     Asamoah Owusu D, Korsah A, Quartey B, Nwolley Jnr. S, Sampah D, Adjepon-Yamoah D, et al. GitHub - Ashesi-Org/Financial-Inclusion-Speech-Dataset: A speech dataset to support financial inclusion [Internet]. GitHub - Ashesi-Org2022;Available from: https://github.com/Ashesi-Org/Financial-Inclusion-Speech-Dataset

15.     Resnik P, Olsen MB, Diab M. The Bible as a Parallel Corpus: Annotating the 'Book of 2000 Tongues.' Comput Hum [Internet] 1999;33:129–53. Available from: https://doi.org/10.1023/A:1001798929185

16.     Mozilla Foundation. Common Voice 17.0: Twi [Internet]. 2023;Available from: https://huggingface.co/datasets/mozilla-foundation/common_voice_12_0/viewer/tw/train

17.     USAID. Language of Instruction Country Profile: Ghana. 2021;1–13. Available from: https://pdf.usaid.gov/pdf_docs/PA00XH27.pdf

18.     Antwi-Boasiako KB, Agyekum K. Globalization , Colonization , and Linguicide : How Ghana is Losing its Local Languages through Radio and Television Broadcast. Int J Humanit Soc Sci 2022;12:142–51.

19.     Agyekum K. Language shift: A case study of Ghana. Sociolinguistic Studies 2009;3:381–403.

20.     Boakye-Yiadom AA, Qin M, Jing R. Research of Automatic Speech Recognition of Asante-Twi Dialect for Translation. In: ACM International Conference Proceeding Series. Association for Computing Machinery; 2021. page 1086–94.

21.     Williams A, Demarco A, Borg C. The Applicability of Wav2Vec2 and Whisper for Low-Resource Maltese ASR [Internet]. In: 2nd Annual Meeting of the ELRA/ISCA SIG on Under-resourced Languages (SIGUL 2023). ISCA: ISCA; 2023. page 39–43.Available from: https://www.isca-archive.org/sigul_2023/williams23_sigul.html

22.     Fuad A, Al-Yahya M. AraConv: Developing an Arabic Task-Oriented Dialogue System Using Multi-Lingual Transformer Model mT5. Applied Sciences (Switzerland) 2022;12.

23.     Abate ST, Tachbelie MY, Schultz T. End-To-End Multilingual Automatic Speech Recognition for Less-Resourced Languages: The Case of Four Ethiopian Languages. In: ICASSP, IEEE International Conference on Acoustics, Speech and Signal Processing - Proceedings. Institute of Electrical and Electronics Engineers Inc.; 2021. page 7013–7.

24.     Chen Y, Liu P, Zhong M, Dou ZY, Wang D, Qiu X, et al. CDEvalSumm: An Empirical Study of Cross-Dataset Evaluation for Neural Summarization Systems. Findings of the Association for Computational Linguistics Findings of ACL: EMNLP 2020 [Internet] 2020 [cited 2025 Jun 22];3679–91. Available from: https://aclanthology.org/2020.findings-emnlp.329/

25.     Das S, Jain A, Durai A, Gabbita S, Vasantharao A, Kotha V. Cross-Dataset Evaluation of Multimodal Neural Networks for Glaucoma Diagnosis. Proceedings - 2022 IEEE 9th International Conference on Data Science and Advanced Analytics, DSAA 2022 2022;

26.     Fried D, Kitaev N, Klein D. Cross-Domain Generalization of Neural Constituency Parsers. ACL 2019 - 57th Annual Meeting of the Association for Computational Linguistics, Proceedings of the Conference [Internet] 2019

[cited 2025 Jun 22];323–30. Available from: https://aclanthology.org/P19-1031/

27.     Fatehi K, Torres Torres M, Kucukyilmaz A. An overview of high-resource automatic speech recognition methods and their empirical evaluation in low-resource environments. Speech Commun [Internet] 2025 [cited 2025 Apr 27];167:103151. Available from: https://www.sciencedirect.com/science/article/pii/S0167639324001225

28.     Gohider N, Basir OA. Recent advancements in automatic disordered speech recognition: A survey paper. Natural Language Processing Journal [Internet] 2024 [cited 2025 Jun 23];9:100110. Available from: https://linkinghub.elsevier.com/retrieve/pii/S294971912400058X

29.     Meyer J, Adelani DI, Casanova E, Öktem A, Weber DWJ, Kabongo S, et al. BibleTTS: a large, high-fidelity, multilingual, and uniquely African speech corpus. arXiv preprint [Internet] 2022;Available from: http://arxiv.org/abs/2207.03546

30.     Kulkarni A, Kulkarni A, Couceiro M, Trancoso I. Unveiling Biases while Embracing Sustainability: Assessing the Dual Challenges of Automatic Speech Recognition Systems. Proceedings of the Annual Conference of the International Speech Communication Association, INTERSPEECH 2024;4628–32.